\documentclass[letterpaper, 10 pt, conference]{ieeeconf}  

\IEEEoverridecommandlockouts                              

\usepackage{graphics} 
\usepackage{epsfig} 
\usepackage{hyperref}
\usepackage{bm}
\usepackage{amsmath, amssymb}

\newcommand{\N}{\ensuremath{\mathcal{N}}}
\renewcommand{\vec}[1]{\ensuremath{\boldsymbol{#1}}}

\title{\LARGE \bf
Low-cost Sensor Glove with Force Feedback for Learning from Demonstrations using Probabilistic Trajectory Representations}

\author{Elmar Rueckert$^{*}$, Rudolf Lioutikov$^{*}$, Roberto Calandra$^{*}$, Marius Schmidt$^{\dag}$, Philipp Beckerle$^{\dag}$ and Jan Peters$^{*,\ddag}$
\thanks{$^{*}$Intelligent Autonomous Systems Lab, Technische Universit\"at Darmstadt,
        Hochschulstr. 10, 64289 Darmstadt, Germany
        {\tt\small \{rueckert, lioutikov, calandra \}@ias.tu-darmstadt.de}}%
\thanks{$^{\dag}$Institute for Mechatronic Systems in Mechanical Engineering, Technische Universit\"at Darmstadt,
        Hochschulstr. 10, 64289 Darmstadt, Germany
        {\tt\small marius.schmidt@stud.tu-darmstadt.de, beckerle@ims.tu-darmstadt.de}}%
\thanks{$^{\ddag}$Robot Learning Group, Max-Planck Institute for Intelligent Systems,
	Tuebingen, Germany
        {\tt\small mail@jan-peters.net}}%
}

\begin{document}


\maketitle
\thispagestyle{empty}
\pagestyle{empty}

\begin{abstract}
Sensor gloves are popular input devices for a large variety of applications 
including health monitoring, control of music instruments, learning sign 
language, dexterous computer interfaces, and teleoperating robot hands 
\cite{sturman1994survey}. Many commercial products as well as low-cost open 
source projects have been developed.\footnote{The gloves project at 
\url{http://theglovesproject.com/data-gloves-overview/} provides an overview.} 
We discuss here how low-cost (approx. $250$ EUROs) sensor gloves with force feedback can be build, 
provide an open source software interface for Matlab and present first results 
in learning object manipulation skills through imitation learning on the 
humanoid robot \textit{iCub}. 
\end{abstract}

\section{INTRODUCTION}

\begin{figure}
\begin{center}
\includegraphics[width=\columnwidth]{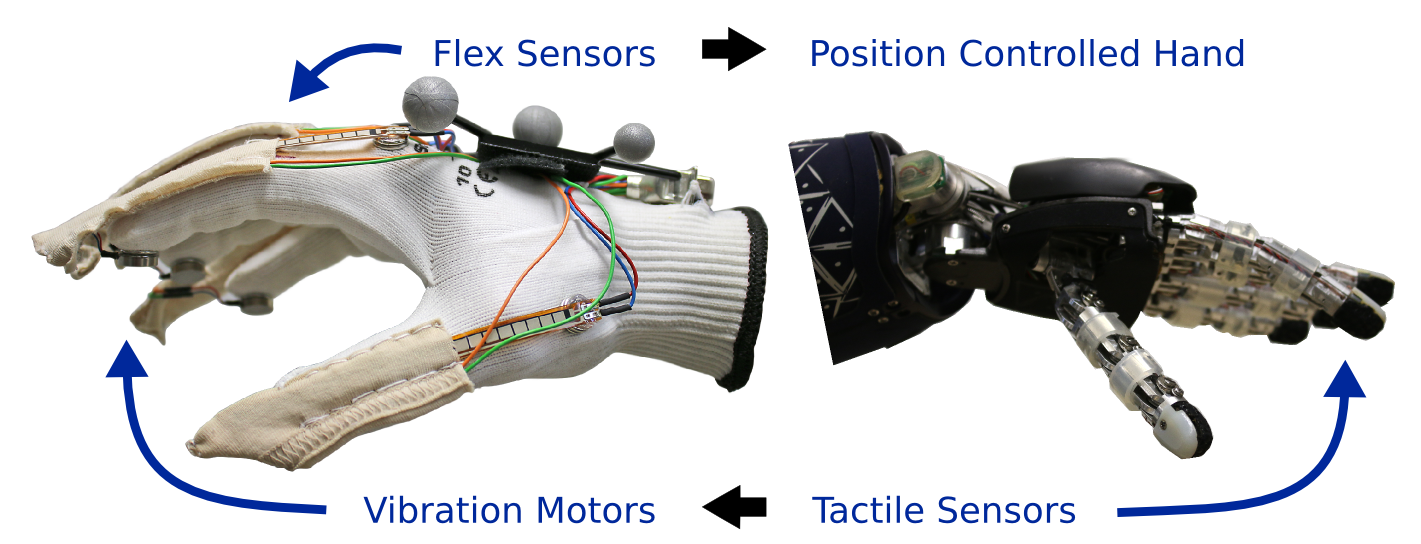}
\end{center}
\caption{A low-cost sensor glove is used to teleoperate a five-finger robot hand. 
The robot hand is equipped with tactile sensors (the \textit{iCub} hand is shown in the picture). 
Tactile information provides force feedback to the teleoperator through activating vibration motors at the glove's fingertips.
\label{fig:gloveAndHand}
}
\vspace{-0.5em}
\end{figure}

In robotics, sensor gloves are widely used for learning manipulation tasks from 
human demonstrations (see e.g., \cite{argall2009survey} for a recent overview). 
From the human operator perspective, active and passive approaches can be 
distinguished. In active approaches, the operator manipulates objects with its 
\textit{own} hand. The robot learns a model through sensing the operator and the 
scene \cite{tung1995automatic, dillmann2000learning, grave2010learning}. The 
demonstrations are platform independent and the learned model can adapt to 
changing environmental conditions. A mapping from human to robot kinematics and 
additional vision systems are needed. 

Alternatively, in passive approaches the operator directly controls the robot 
hand through an instantaneous mapping from sensor glove measurements to control 
actions \cite{fischer1998learning, henriquesplanning}. This has the advantages  
that the human teleoperator can adapt for the limited capabilities of the robot 
(compared to humans) or to compensate for kinematic mapping errors. Moreover, 
joint and contact forces can be recored at the robot side and used to train 
inverse dynamics control methods. Additional force feedback at the sensor glove 
provides important information about contact forces during grasping.  The 
drawback of passive approaches is that the demonstrations are platform specific 
and rapid learning methods are needed for teaching a large variety of robots new 
skills. We follow this line of research here. 

Most models of demonstrations model single trajectories, either directly or 
indirectly (see e.g., \cite{Schaal_ISRR_2004} and \cite{kober_MACH_2011}). A 
tracking controller with fixed or pre-tuned gains is used to follow the model 
prediction. However, for a safe operation of robots in everyday environments low 
gain control strategies are needed. Such compliant control approaches can be 
computed from the variance of the distribution over \textit{multiple} 
trajectories. In probabilistic movement representations optimal control gains 
can be inferred that reproduce the demonstrations \cite{Paraschos_NIPS_2013a} or 
solve optimal control problems \cite{Rueckert_PMP_Frontiers}. Here, we follow a 
simpler approach. As in probabilistic movement representations, a distribution 
over trajectories is learned using Bayesian linear regression with Gaussian 
features. Instead of inferring an optimal control law (which requires a forward 
model) we exploit a built-in low-level impedance controller to track the mean 
of the learned trajectory distribution. Variance dependent adaptive control 
gains approaches are part of future research. 




\section{METHODS}

\begin{figure*}[!t]
\begin{center}
\includegraphics[width=2\columnwidth]{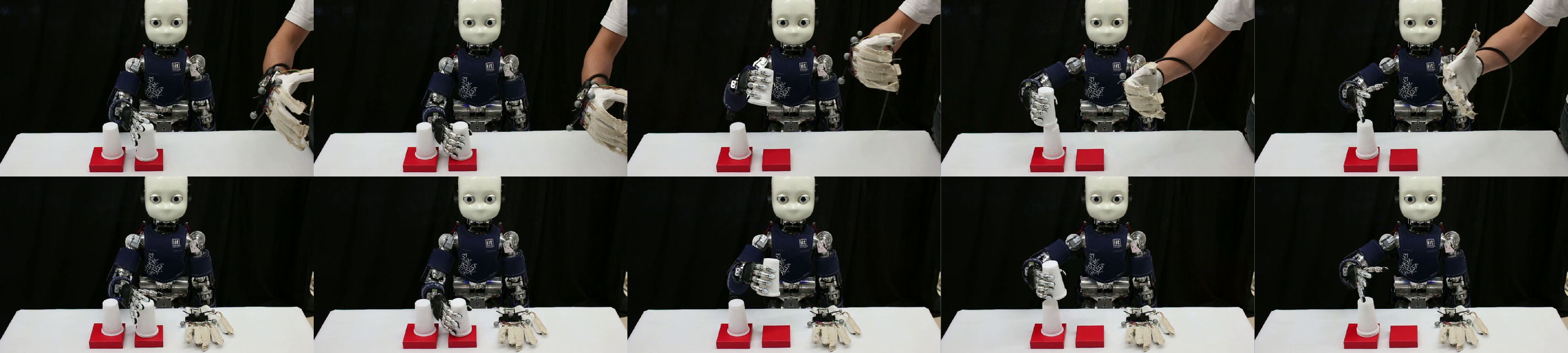}
\end{center}
\caption{Learning a cup stacking task. Top row: Demonstration through teleoperation. The robot arm is controlled using an approximation of the inverse kinematics following an end-effector trajectory provided by a motion capturing system. The sensor glove provides desired finger joint angles.   
Bottom row: Autonomous reproduction. An impedance controller is used to track the maximum a posteriori estimate of the trained probabilistic movement primitives.  
\label{fig:cupStackingSequence}
}
\vspace{-0.5em}
\end{figure*}

\subsection{Sensor Glove Hardware}

Our glove is based in the \textit{animatronic robot hand project2}
\footnote{The mechanics and the electronics are detailed here: \url{http://goo.gl/TqPYdK}.} \cite{de2014developing}, 
where we added force feedback. 
Finger motion data is acquired by $4.5$ inch flex sensors measuring the bending of 
each finger, see Figure \ref{fig:gloveAndHand}. The range of resistance is linearly mapped to the range of 
motions. To introduce force feedback, VPM2 coin vibration motors are placed at 
the fingertips. Those are controlled by mapping the tactile sensor readings of 
the robot hand proportionally to motor pulse width modulation (PWM) values. 
All sensors and motors are connected to an arduino board (mega 2560) that processes the data.

\subsection{Glove Software Interfaces}

The arduino board implements a serial communication interface ($115200$ Baud) 
and communicates with a computer through USB. In the current implementation, the 
communication protocol streams the flex sensor readings at a rate of $350$Hz 
(with a resolution of two bytes per measurement). The sensor values can be 
accessed through a callback event raised by a Matlab \textit{mex} function. 
Force feedback values can be send to the glove in form of a string command 
representing pulse width modulation (PWM) values $\in [0,255]$.\footnote{Matlab code for the glove: \url{http://tinyurl.com/nf96nr7}, 
a Java interface: \url{http://tinyurl.com/o4hjumq}.}

\subsection{Learning from Demonstrations}

We denote a single demonstration as sequence of $T$ state vectors $\vec \tau = \vec y_{1:T}$. 
The state vector is modeled in a 
linear basis function model assuming zero mean Gaussian noise where 
$\vec y_t = \vec \Phi_t \; \vec w + \vec \epsilon_y$ 
with $\vec \epsilon_y \sim \N\left(\vec \epsilon_y \, | \, \vec 0, \vec \Sigma_y \right)$. 
The matrix $\vec \Phi_t$ denotes the (extended) feature matrix using Gaussian basis functions \cite{Rueckert_ICRA14LMProMPsFinal}.

To model a distribution over multiple trajectories we introduce the prior distribution $p(\vec w) = \N\left(\vec w\middle| \vec \mu_w, \vec \Sigma_w \right)$. 
The mean $\vec \mu_w$ and the covariance matrix $\vec \Sigma_w$ can be learned from data 
by maximum likelihood using the Expectation Maximization (EM) algorithm \cite{Dempster1977}.  
A simpler solution that works well in practice is to compute first the most likely estimate of $\vec w^{[i]}$ for each trajectory $\vec \tau^{[i]}$ independently 
(where the index $i$ denotes the $i$-th demonstration).   
Given a trajectory $\vec \tau^{[i]}$, the corresponding 
weight vectors $\vec w^{[i]}$ can be estimated by a straight forward least squares estimate
 $\vec w^{[i]} = \left(\vec \Phi_{1:T}^T \, \vec \Phi_{1:T} + \lambda \, \vec I\right)^{-1} \, \vec \Phi_{1:T}^T \; \vec \tau^{[i]}$. 
Subsequently, the mean and the covariance of $p(\vec w)$ can be estimated 
by the sample mean and sample covariance of the $\vec w^{[i]}\,$'s. 

After learning, the probability of trajectory $\vec \tau$ given the feature vector $\vec w$ reads  
$p(\vec \tau \, | \, \vec w) = \prod_{t=1}^T \N\left(\vec y_t \middle| \vec \Phi_{t} \, \vec w, \vec \Sigma_y \right)$.  
For more details we refer to \cite{Paraschos_NIPS_2013a, Rueckert_ICRA14LMProMPsFinal}. 
Demo Matlab code was made open source.\footnote{Matlab code of the model: \url{http://tinyurl.com/qj4joxz}.}


\section{Cup Stacking Task on the iCub}

The sensor glove was used to teach the humanoid robot \textit{iCub} how to stack two 
plastic cups as depicted in Figure \ref{fig:cupStackingSequence}. Prior to the 
demonstrations, we recorded the minima and maxima of the five flex sensors when the 
teleoperator flexed and spread its fingers. These values were used to in a 
linear mapping from flex readings to the robot finger joint angles (the joint 
angle extrema of the robot fingers were manually determined). The arm joints 
were controlled through an inverse kinematics approximation given the Cartesian 
coordinate and orientation of the operator's wrist (measured in a motion 
capturing system). A built-in low-level impedance controller with manually 
tuned gains was used to track the desired joint angles during teleoperation or 
after training in the autonomous execution phase. The control rate was $200$Hz 
and the movement duration was set to $15$ seconds. 

\begin{figure}
\begin{center}
\includegraphics[width=0.8\columnwidth]{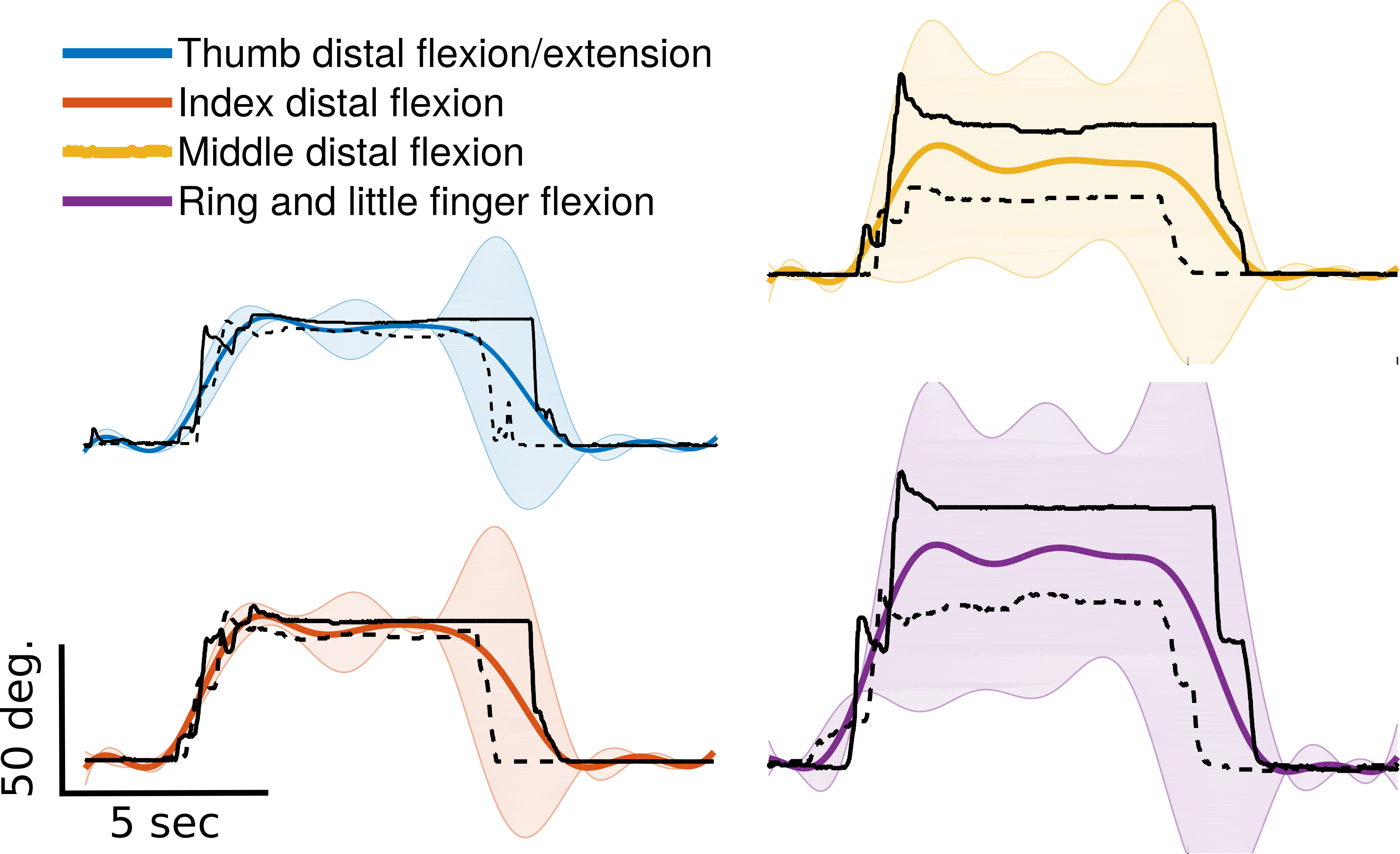}
\end{center}
\caption{Illustration of the joint encoder readings of four distal finger joints for two demonstrations (solid and dashed black lines). 
The little and the ring finger are coupled in the five finger hand. 
The model mean is denoted by the smooth colored lines. It is used in the autonomous reproduction phase in a tracking controller. 
The standard deviation is denoted by the shaded area.
\label{fig:CupStackingTrajAndModel}
}
\vspace{-0.5em}
\end{figure}

Two demonstrations were used for training the probabilistic model.  In total, a 
model of $13$ joints was learned. For the distal finger joints the 
demonstrations and the learned model are shown in Figure 
\ref{fig:CupStackingTrajAndModel}. The model filters the demonstrations through 
averaging over the demonstrations. Note that the force feedback was helpful for 
grasping the cup without deforming it. This needs to be further evaluated. A 
successful autonomous execution tracking the mean of the learned distribution is 
shown in the bottom row in Figure \ref{fig:cupStackingSequence}.


\section{CONCLUSIONS}

A low-cost sensor glove with force feedback was presented. We used it in first experiments for 
imitation learning in a cup stacking task in the humanoid robot \textit{iCub}. 
We developed open source code for the sensor glove and the probabilistic model. 
Future hardware extensions include an inertial measurement unit (to replace the motion capturing system) 
and additional flex sensors to measure thumb adduction/abduction and proximal joint motions. 
We also plan to exploit the learned variance 
in the probabilistic model for adaptive control gains.



\section*{ACKNOWLEDGMENT}

The research leading to these results has received funding from the European 
Community's Seventh Framework Programme (FP7/2007-2013) under grant agreement 
No. 600716 (CoDyCo). The authors would like to 
acknowledge Alexander Burkl, Chao Su, and Ji-Ung Lee for their 
assistance during the glove development.

\bibliographystyle{unsrt}
\bibliography{refs}

\end{document}